\documentclass[10pt,conference]{ieeeconf} % uncomment this

\usepackage{setspace}
\usepackage{psfrag,subfigure}
\usepackage{amssymb, amsmath, cite}
\usepackage{color}
\usepackage{verbatim}
\usepackage{mathrsfs}
\usepackage{graphicx}

\usepackage[top=1in, bottom=0.7in, left=0.7in, right=0.7in]{geometry}
\usepackage{mathtools}
\usepackage{url}
\usepackage{pdfpages}

\IEEEoverridecommandlockouts % uncomment this
\newcommand{\RR}{{\mathbb R}}
\newcommand{\TT}{{\mathbb T}}

\newcommand{\cB}{{\mathcal B}}

\newcommand{\cP}{{\mathcal P}}

\newcommand{\cS}{{\mathcal S}}

\newcommand{\conv}{\textup{co}}

\title{\Large \bf Safe and Robust Robot Maneuvers Based on Reach Control}
 
\author{Marijan Vukosavljev, Ivo Jansen, Mireille E. Broucke, and Angela P. Schoellig% <-this % stops a space
\thanks{Marijan Vukosavljev and Mireille E. Broucke are with the Dept. of Electrical and Computer Engineering, University of Toronto, Canada (e-mails: mario.vukosavljev@mail.utoronto.ca, broucke@control.utoronto.ca). Ivo Jansen was a visiting student from the Dept. of Mechanical Engineering, Eindhoven University of Technology, the Netherlands (email: i.h.m.jansen@student.tue.nl). Angela P. Schoellig is with the University of Toronto Institute for Aerospace Studies (UTIAS), Canada (email: schoellig@utias.utoronto.ca). Supported by the Natural Sciences and Engineering Research Council of Canada (NSERC). }%
\thanks{Associated video at: http://tiny.cc/quadrotorRCPx.}
}

\begin{document}

%\linespread{0.9} % in case tight for space!

\maketitle

\begin{abstract}

In this paper, we investigate the synthesis of piecewise affine feedback controllers to address the problem of safe and robust controller design in robotics based on high-level controls specifications. The methodology is based on formulating the problem as a collection of reach control problems on a polytopic state space. Reach control has so far only been developed in theory and has not been tested experimentally on a real system before. Using a quadrocopter as our experimental platform, we show that these theoretical tools can achieve fast, albeit safe and robust maneuvers. In contrast to most traditional control techniques, the reach control approach  does not require a predefined open-loop reference trajectory or spacial path. Experimental results on a quadrocopter show the effectiveness and robustness of this control approach. In a proof-of-concept demonstration, the reach controller is implemented in one translational direction while the other degrees of freedom are stabilized by separate controllers. 
\end{abstract}

\section{Introduction}
\label{sec:intro}

This paper proposes a novel framework for control of complex robotic maneuvers that simultaneously impose requirements of safety, fast response, and a desired sequence of events. We apply the framework to a simple side-to-side maneuver on a quadrocopter in order to expose the main features of the framework. The framework is based on using hybrid systems, event-based switching, and solving a collection of so-called reach control problems (RCP). The RCP has an extensive theoretical development, see \cite{MEB10, BG14, HVS04, HVS06, RB06}, but it has been completely lacking in experimental validation. The primary goal of this paper is to illustrate, for the first time  on a real system, the various strengths offered by the reach control approach. 

Because our main application is quadrocopter maneuvering, we give an overview of current approaches, particularly placing our reach control approach within this literature. There are two predominant methods for control of quadrocopter maneuvers: timed trajectory tracking and path following. In timed trajectory tracking, an open-loop reference trajectory as a function of time is predefined. Then a controller that stabilizes the system to the trajectory is designed. For example, in~\cite{LUP2014}, an impressive collection of aggressive quadrocopter maneuvers is featured using this approach. Timed trajectory tracking is the most common method in the literature \cite{LUP2014, SCH2011, DandOPT}. On the other hand, difficulties arise in finding the open-loop reference trajectory. For example, in \cite{DandOPT} it  is first geometrically specified using splines, and then time parameterized so that the resulting reference trajectory is feasible. Also, while timed trajectory tracking can provide high-performance maneuvers, due to the open-loop nature of the reference signals, any additional unaccounted disturbances can quickly deteriorate performance.

In path following, a 3D spatial (untimed) path is specified in output space. Then an output stabilization method is used to keep the system on the path. Recent applications of path following to quadrocopter maneuvering include \cite{KUM2012,PAS2013,ROZ2012}. The benefit of this method compared to timed trajectory tracking is better control of transients: when the system deviates from the path, it must only steer back to the path rather than to a specific point in time. The difficulty in this method is again finding paths feasible for the dynamics and constraints.

In our approach, it is not required to generate a feasible timed trajectory or path. Rather, control specifications are given that restrict the dynamics to a feasible region of the state space. Further, these specifications inform on how the state must evolve in that region. The region is  then partitioned into smaller regions (in our case simplices) and a feedback controller is designed for each region to guarantee correct evolution of the state. As such, our method has several significant advantages: 1) we bypass the construction of a reference trajectory or path; 2) we obtain feedback controllers, not open-loop controls; 3) we obtain controllers for the entire feasible region of operation, not only a neighborhood of a path; 4) we explicitly account for safety constraints and actuator limits. The main difficulty of our method (as it is implemented right now) is its application to high-dimensional systems, where the required state-space partitioning becomes more involved. In the future, advanced computational tools may be adopted from other fields to address this problem.
   On the other hand, this paper shows that high-order models can be reduced to simpler models with no degradation of performance. For related methods to our approach see \cite{GRIZ14,TOM2011}.  Reach control has never been applied on a real system before.  This work presents a proof-of-concept of its practical applicability.  

\section{Methodology}
\label{sec:rcpconcept}

In this section, we briefly outline the reach control methodology, which allows us to define high-level controls specifications, and results in safe and robust system behavior.
\subsection{System Model and Control Specifications}
We assume that the system is modeled as a finite-dimensional, dynamical system, with possibly nonlinear dynamics:
\begin{equation} \label{eq:sysnonlin}
\dot{\tilde{s}} = \tilde{f}(\tilde{s},\tilde{u}),
\end{equation}
\noindent where $\tilde{s} \in \RR^n$ is the state and $\tilde{u} \in \RR^{n_u}$ is the control input. 

The objective of the controller is defined in terms of high-level safety and event specifications. In particular, the approach can handle the following types of specifications: 

\paragraph{Safety and Liveness} Safety and liveness constraints define the region in the state space that the system is allowed to visit. Safety constraints limit the dynamics to a safe regime of operation. Liveness constraints enforce fast, lively response. Our framework requires that the feasible region be a polytope $\cP$, as shown in Figure \ref{fig:polytope}.
%We introduce two types of constraints: safety and liveness. The constraints involving safety typically encode how to avoid collisions. The constraints involving liveness encode which configurations of the system lead to sluggish execution of the desired system behavior. Together, both constraints define the operational region of the state space where the states of the modeled robotic system must remain indefinitely. For our methodology, this operational region must be described as a polytope on the relevant states \textcolor{red}{(define mathematically a non-convex polytope?)(Notice that other works define these regions differently, see for example control lyapunov barrier functions literature, so do a comparison?)}. For complex systems, even an approximate computation of this operational region quickly becomes intractable, see also Chapter 5 of \cite{SPO05}. 

\paragraph{Desired Temporal Sequence} This specification describes the overall sequence of events or the overall  motion of the system. A set of target states to be reached by the system must be defined. For more complex maneuvers, a sequence of target sets may be defined, and can be formalized by using automata from discrete event systems \cite{RAMADGE}. For example, Figures  \ref{fig:polytope} and \ref{fig:triangulation}  show that the system must traverse the polytope $\cP$ clockwise by moving through a sequence of triangles $\cS_i$, $i = 1, 2, \ldots$ .
%Furthermore, each discrete mode of this automaton, which is characterized by reaching a particular target set, may have a different polytope corresponding to safety and liveness constraints. See 
%Next we introduce the desired temporal sequence, which describes the overall sequence of events or motion of the system. To assist in describing this, automata from discrete event systems can be used, especially if the underlying robotic system has hybrid dynamics \textcolor{red}{(formally introduce automata notation?)}. For each discrete mode of operation, a set of target states in the system's state space must be defined; upon reaching a target state, a discrete event occurs to proceed the system to the next discrete mode that contains a new set of target states, and so on. In fact, for each discrete mode, one may use a different polytope to describe the operational region for that mode. In this way, complex behavior can be broken down into manageable pieces.

The specifications above can handle  complex tasks, but can be computationally difficult for high-dimensional systems; a higher-level control architecture may help to decouple the complexity and will be illustrated in our quadrocopter example.
%The specifications above allow to encode very complex tasks, but can be daunting to determine precisely, especially on high-dimensional systems. The quadrocopter example in this paper will show one way of complexity reduction, but in general each specific problem may benefit from some prior ingenuity to cast the problem into a lower dimensional form before applying this methodology \textcolor{red}{(this is a very important point, my crane paper also had some ``ingenuity", namely the output rcp using trial and error, so reference that too?)}.

To summarize, the control specifications require determining a polytope, which describes the allowable states and a sequence of target sets. Given this data, the objective is to find a controller that ensures the states of the robotic system  remain in the polytope while reaching the correct sequence of target sets. For a related example on complex control specifications in the context of reach control, see \cite{VUK14}.

\subsection{Triangulation of the Polytope}
%To drive any initial state starting in $\cP$ to a target state while remaining in $\cP$ can be difficult. To systematize this, we partition $\cP$ into a set of regions and specify a controller on each region. Since $\cP$ may be a non-convex polytope, we employ a triangulation of $\cP$ into simplices, see Figure \ref{fig:triangulation}. Informally,  the polytope is covered fully by non-intersecting, $n$-dimensional triangles.
To drive any initial state starting in $\cP$ to a target state while remaining in $\cP$ can be difficult. To systematize the design, we triangulate $\cP$ into a set of simplices and we specify a controller on each simplex, see Figure \ref{fig:triangulation}. Informally, the polytope is partitioned into triangles. 
 
Formally, an $n$-dimensional {\em simplex}, $\cS \coloneqq \conv\{v_0, \ldots, v_n\}$, is the convex hull of $(n+1)$ affinely 
independent points in $\RR^n$; it is the generalization of a triangle.
A \emph{facet} of a simplex is a boundary face of dimension $(n-1)$. 
A {\em triangulation} is a partition of a set $\cP \subset \RR^n$ into $n_p$~simplices and is 
denoted as $\TT = \{\cS_1, \ldots, \cS_{n_p}\}$, see \cite{LEE}. Then $\TT$ satisfies the properties: 
\begin{enumerate}
\item[(i)]   
$\TT = \cS_1 \cup \ldots \cup \cS_{n_p}$ and
\item[(ii)]  
$\cS_i \cap \cS_j, i \neq j,$ is a lower-dimensional simplex of both $\cS_i$ and $\cS_j$ or the 
empty set for all $i, j \in \{1, \ldots, n_p\} $.
\end{enumerate}

Once a triangulation of $\cP$ has been specified, the next step of the design is to identify 
a sequence of simplices to be visited in order  to be able to
reach the target states. 

Using the sequence of simplices, \emph{exit facets} for each simplex are designated. 
The trajectories starting in the given simplex may only exit the simplex through the exit facets, while the remaining facets act as \emph{restricted facets}. 

On each simplex, it is assumed that the dynamics of the system are affine, having the form 
\begin{equation} \label{eq:affinesys}
\dot{s} = As + Bu + a,
\end{equation} 
\noindent where $s \in \RR^n$, $u \in \RR^{n_u}$, and the matrices have appropriate dimensions. If the dynamics \eqref{eq:sysnonlin} are nonlinear, they may be linearized about some point in the simplex to yield the form~\eqref{eq:affinesys}. 

\subsection{Control Design via the Reach Control Problem} \label{sec:RCPmath}

Finally, controllers are designed for each simplex based on the {\em reach control problem} (RCP). This method ensures that closed-loop trajectories flow through the designated exit facets without crossing the restricted facets.
The reach control problem formulation, its theoretical developments, and conditions for solvability are 
discussed in \cite{MEB10, BG14, HVS04, HVS06, RB06}. 
%The main purpose of RCP is to guide trajectories of a dynamical system through a specific region of the state space defined by safety and performance requirements. Ultimately, the trajectories should reach a target set of states in finite time without violating the boundaries of the specified region. 
%

%\begin{comment}
\begin{figure}
%2D simplex, cone(v0), facets, normal vectors, vertices, xi
\centering
\includegraphics[width=0.5\linewidth]{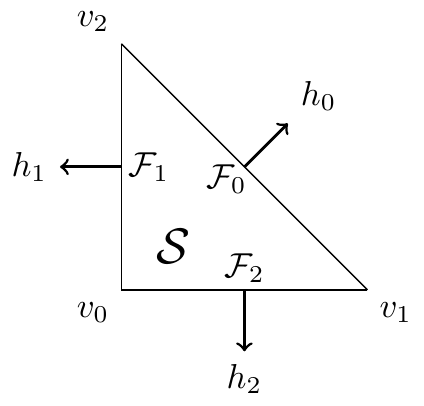}%
%\begin{comment}
%\begin{tikzpicture}[scale=0.5,every node/.style={scale=1}]
%\pgfmathsetmacro{\X}{5}
%\coordinate (v0) at (0,0);
%\coordinate (v1) at (\X,0);
%\coordinate (v2) at (0,\X);
%\draw (v0) -- (v1) node[below left] at (v0) {$v_0$};
%\draw (v1) -- (v2) node[below right] at (v1) {$v_1$};
%\draw (v0) -- (v2) node[above left] at (v2) {$v_2$};
%\draw [->, thick] (\X/2,0) -- (\X/2,-\X/4) node[below] at (\X/2,-\X/4) {$h_2$};
%\draw [->, thick] (0,\X/2) -- (-\X/4,\X/2) node[left] at (-\X/4,\X/2) {$h_1$};
%\draw [->, thick] (45:\X*0.707) -- (45:\X*0.957) node[above right] at (45:\X*0.957) {$h_0$};
%\node[right] at (-0.1,\X/2) {$\mathcal{F}_1$};
%\node[above] at (\X/2,-.1) {$\mathcal{F}_2$};
%\node[below left] at (45:\X*0.79) {$\mathcal{F}_0$};
%\node at (40:\X*2.2/8) {{\Large $\mathcal{S}$}};
%\end{tikzpicture}
%\end{comment}
\caption{Two-dimensional simplex and related terminology.} \label{fig:simplex}
\vspace{-2mm}
\end{figure}
%
%\end{comment}

Below we summarize the procedure for determining a controller over an arbitrary simplex in our triangulation; see Figure \ref{fig:simplex} for a 2D example. 

We define $\mathcal{I}_n = \{0,1, \ldots, n \}$ to be the index set for the $(n+1)$ vertices of the simplex. The coordinates of each vertex are denoted as $v_i \in \RR^n$, $i \in \mathcal{I}_n$. For each of the vertices, $v_i$, we must pick a corresponding $u_i \in \RR^{n_u}$. 
%In the sequel, we use the term \emph{control value} to refer to the control assignments $u_i$ specifically at the vertices $v_i$ of $\cP$. 
Each facet, $\mathcal{F}_i$, of the simplex is indexed by the vertex index that it does {\em not} contain, and each facet has an associated normal vector $h_i$ to describe its orientation, see Figure \ref{fig:simplex}. We assume that there is at least one restricted facet, and index the restricted facets using $\mathcal{I}_r \subset \mathcal{I}_n$. 

To solve RCP on a given simplex, we must select controls $u_i$ at the vertices $v_i$ to satisfy the so-called \emph{invariance conditions} \cite{RB06}; 
that is,
\begin{equation}
\label{eq:invconds}
(\forall i \in  \mathcal{I}_n)(\forall j \in \mathcal{I}_r \backslash \{i\}) \;\;
h_j \cdot (Av_i + Bu_i + a) \leq 0 .
\end{equation}
\noindent This condition encodes that the velocity vector at each vertex points in the right direction so that trajectories leave the simplex through an exit facet while avoiding crossing the restricted facets.
% an appropriately chosen control at each vertex, the velocity vector at each vertex points in the ``right direction" so that trajectories leave the simplex through an exit facet while avoiding crossing the restricted facets. 
The feasibility of the inequalities in \eqref{eq:invconds} can be easily checked numerically via a linear program. If they are not feasible, then RCP is not solvable and the choice of restricted and exit facets must be modified. If they are feasible, then for a feasible choice of $u_i$, $i \in \mathcal{I}_n$, it can be shown that %\textcolor{red}{by utilizing convexity} 
one can construct an affine feedback to be used over the entire simplex, see \cite{HVS04}. The affine feedback controller has the form
\begin{equation} \label{eq:affinefblaw}
u(t)=K_c s(t)+g_c,
\end{equation}
\noindent where $K_c$ and $g_c$ are obtained using
\begin{equation}
\label{eq:affinefb}
\begin{bmatrix} K_c^{\top} \\ g_c ^{\top} \end{bmatrix} = \begin{bmatrix} v_0^{\top} & 1 \\ \vdots & \vdots \\ v_n^{\top} & 1 \end{bmatrix}^{-1} \begin{bmatrix} u_0^{\top} \\  \vdots \\ u_n^{\top} \end{bmatrix}.
\end{equation}

The final step is to check that the closed-loop system, $\dot{s} = (A+B K_c)s+(a+B g_c)$, contains no equilibrium in the simplex. If so, then RCP is solved over the simplex. For a more detailed discussion on the design of RCP controllers, see \cite{AB13B}.

In summary, the resulting control law over $\cP$ is a piecewise affine feedback with switching between controllers occurring at the boundaries between contiguous simplices. 

\section{Application to a Quadrocopter Maneuver}
We follow the methodology described earlier to design a controller for executing a simple side-to-side quadrocopter maneuver. Due to the complexity of the quadrocopter system, our overall control strategy relies on a standard quadrocopter control architecture, depicted in Figure \ref{fig:architecture}, that decouples the various degrees of freedom, see Section \ref{sec:model}. 
%With a little analysis, we show that this control architecture (luckily) allows us then to employ the methodology described earlier on only one degree of freedom. 
%The focus of this paper is the design of the Reach Controller which is responsible for generating a desired pitch angle to be input for the onboard controller. 
%One degree of freedom, $\theta_d$, is responsible for controlling the side-to-side motion aspect of the maneuver and is the main point of focus in this paper. 
One degree of freedom, corresponding to the design of the Reach Controller in Figure \ref{fig:architecture}, is responsible for controlling the side-to-side motion aspect of the maneuver and is the main point of focus in this paper.
The remainder of the control architecture is standard and ensures that the quadrocopter executes the side-to-side motion while stabilizing the remaining degrees of freedom. 

\subsection{Quadrocopter Model}
\label{sec:model}

\begin{figure}[t]%
\centering%
\includegraphics[width=0.8\linewidth]{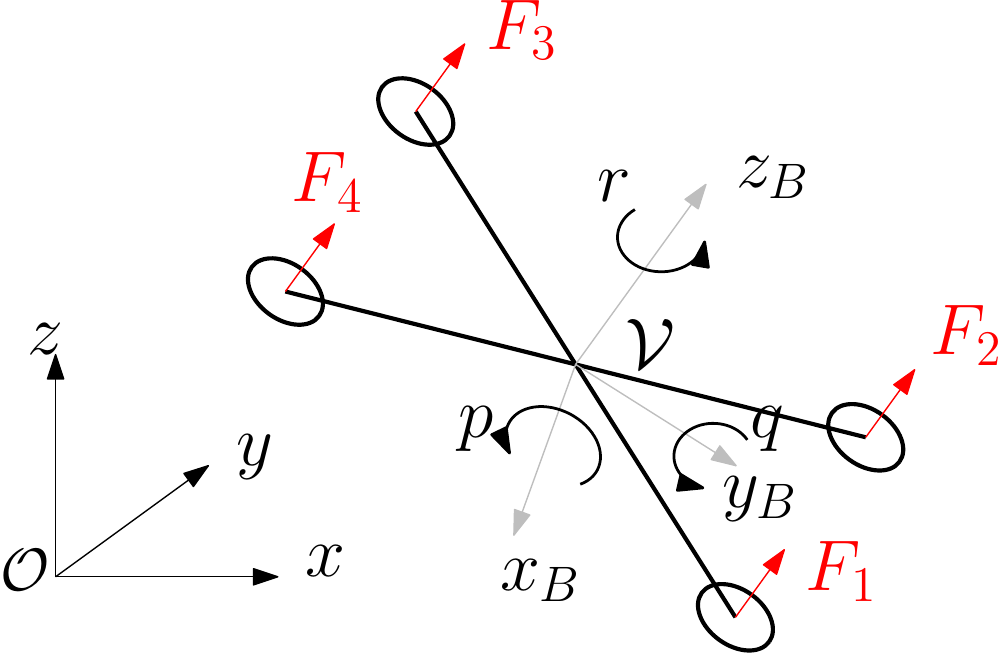}%
\caption{The inertial and quadrocopter body-fixed frames $\mathcal{O}$ and $\mathcal{V}$. The quadrocopter is actuated by varying the thrusts $F_i$, $i \in \{1,2,3,4\},$ produced by each motor. This results in changes to its body rotation rates, $(p,q,r)$, and vertical acceleration, which then causes a change to the quadrocopter's position and attitude.}%
\vspace{-2mm}
\label{fig:quadModel}%
\end{figure}

The quadrocopter model is ubiquitous in the literature; see, for example, \cite{SCH2011,LUP2014} or Chapter 4 of \cite{SCH2014};
we refer the reader to those references for details. The vehicle dynamics are described by six degrees of freedom and are nonlinear. 
The translational position $(x,y,z)$ is measured in the inertial coordinate system $\mathcal{O}$ as shown in 
Figure \ref{fig:quadModel}. The vehicle attitude is defined by the body-fixed frame $\mathcal{V}$ and is represented 
by the $ZYX$-Euler angles, yaw, pitch, and roll, $(\psi, \theta, \phi)$. The full state of the vehicle additionally 
includes the translational and rotational velocities of the body frame, $(\dot{x}, \dot{y},\dot{z})$ represented
in $\mathcal{O}$ and $(p, q, r)$ represented in $\mathcal{V}$, respectively.   

\begin{figure}[t]%
\centering%
\includegraphics[width=1\linewidth]{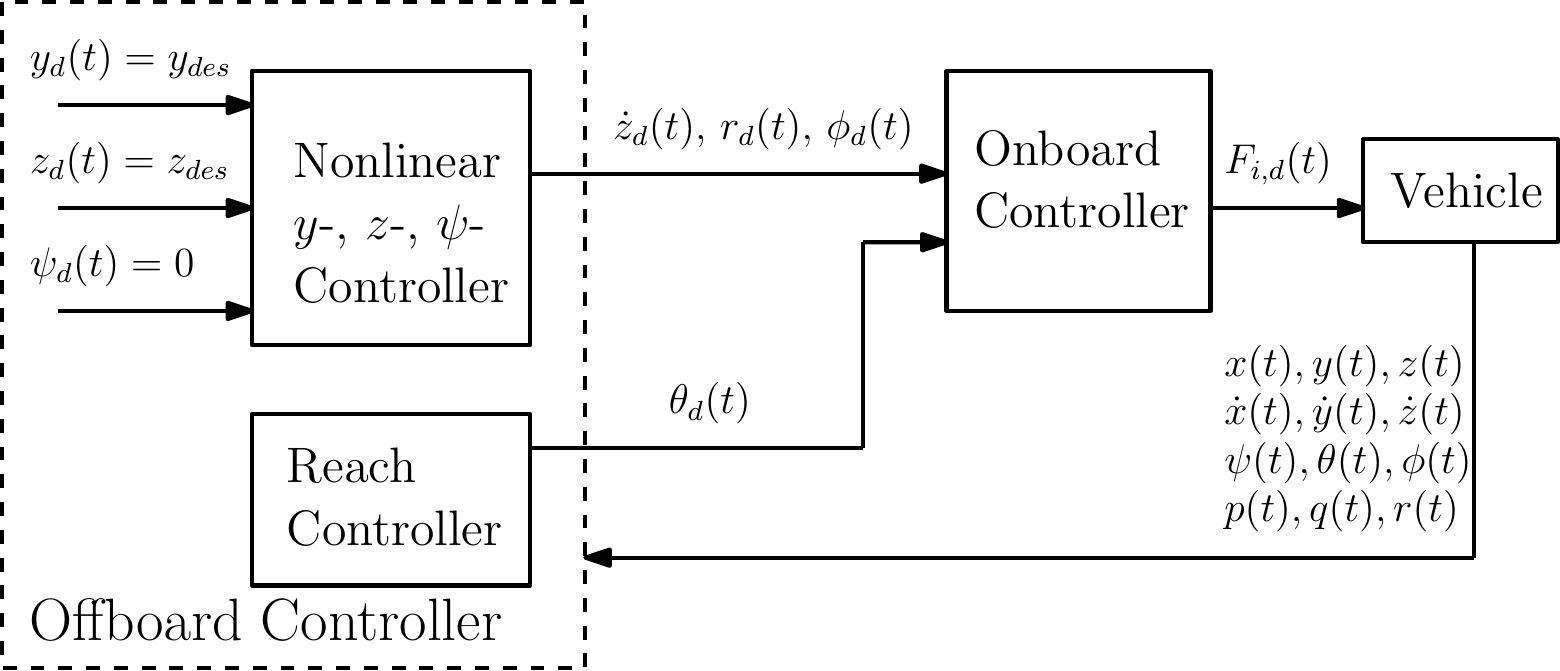}%
\caption{The control architecture.}%
\label{fig:architecture}%
\end{figure}

In our control architecture (Figure \ref{fig:architecture}), we assume that the full state of the vehicle is measured. An onboard controller takes the desired pitch angle $\theta_d$, roll angle $\phi_d$, angular body velocity around the body's $z$-axis $r_d$, and vertical velocity of the vehicle $\dot{z}_d$ as inputs and calculates the required motor forces  $F_{i,d}$, $i \in \{1, 2, 3, 4\}$. In our experiment, the onboard controller is an unmodifiable blackbox. In the offboard controller, a standard, nonlinear tracking controller (as, for example, proposed in \cite{LUP2014}) is used for stabilizing the $y$- and $z$-coordinates of the vehicle as well as the yaw. We use the Reach Controller for the $x$-direction.

%The focus of this paper is the design of the Reach Controller which is responsible for generating a desired pitch angle to be input for the onboard controller. 
%Below we provide an analysis that shows how to apply our methodology in a low dimensional way, with $n=2$.

We assume that the nonlinear controller successfully stabilizes the vehicle at $y_d(t) = y_{des}$, $z_d(t) = z_{des}$, and $\psi_d(t) = 0$, $y_{des}, z_{des} \in \RR$, and provides the onboard controller inputs $\phi_d(t)$,  $r_d(t)$, and $\dot{z}_d(t)$. 
The equations governing the $x$- and $z$-motion of the vehicle are then given by
\begin{align} 
\label{eq:ddotxsimple} \ddot{x}(t) &= f(t) \sin\left( \theta(t) \right)\\
\label{eq:ddotzsimple} \ddot{z}(t) &= f(t) \cos \left( \theta(t) \right) - g ,
\end{align}
\noindent where $g$ is the gravitational constant and $f(t)$ is the collective thrust normalized by the vehicle mass $m$, 
\begin{equation}
\label{eq:sumrotor}
f(t) = \frac{1}{m} \sum_{i=1}^4 F_i(t)  ,
\end{equation}
\noindent with motor forces $F_i$, $i \in \{1, 2, 3, 4\}$, see Figures \ref{fig:quadModel} and \ref{fig:architecture}.

Since $z(t) = z_{des}$ implies $\ddot{z}(t) = 0$, equation \eqref{eq:ddotzsimple} gives $f(t) = g / \cos \left( \theta(t) \right)$, and with \eqref{eq:ddotxsimple} we have
\begin{equation} \label{eq:ddotxDI}
\ddot{x}(t) = g \tan \left( \theta(t) \right) := u(t) \; \Leftrightarrow \; \theta(t) = \tan^{-1}\left( \frac{u(t)}{g} \right) .
\end{equation}

To summarize this analysis, if we can define an $\ddot{x}$ profile, then with $u(t) = \ddot{x}(t)$ and \eqref{eq:ddotxDI} this will produce a desired pitch angle signal $\theta_d(t)$ to be input for the onboard controller (see again Figure \ref{fig:architecture}). The signal $u(t)$ will be constructed via the reach control methodology outlined earlier
and this will yield a feedback $u(t) = u(x(t),\dot{x}(t))$. 

This approach can be extended to three dimensions: similarly as above, each direction can be formulated as a single- or double-integrator, linear system. The challenge is in partitioning the state space. A recent method for complexity reduction involves working in output space \cite{KRO16}.
%\textcolor{red}{We also mention that for an extension in the $y$ and $z$ directions, one must relax $y_d(t) = y_{des}$, $z_d(t) = z_{des}$. A similar analysis would lead one to define new control variables $v$ and $w$ respectively. Overall, the onboard controller inputs $\theta_d(t)$, $\phi_d(t)$, and $\dot{z}_d(t)$ would be determined by $u$, $v$, and $w$. The corresponding control specifications and RCP design for $u$, $v$, and $w$ would also be higher dimensional. This will be demonstrated in future work.} \textcolor{blue}{This is my attempt to highlight how to extend to 2D and 3D but should be made shorter}

%The power of our methodology for determining $u(t)$ will become apparent from our experimental results: obviously no system can attain perfect tracking, as assumed earlier for the nonlinear controller. Yet the results suggest that this construction for $u(t)$ is not sensitive to imperfect tracking can still execute the desired maneuver correctly. \textcolor{red}{emphasize that we do not have any theoretical results on how bad tracking can be before everything falls apart?}
%The details are given in Section \ref{sec:rcpconcept}. 

\subsection{Control Specifications}
\label{sec:specs}

\begin{figure}[t]%
\centering%
\includegraphics[width=0.9\linewidth, height=0.6\linewidth]{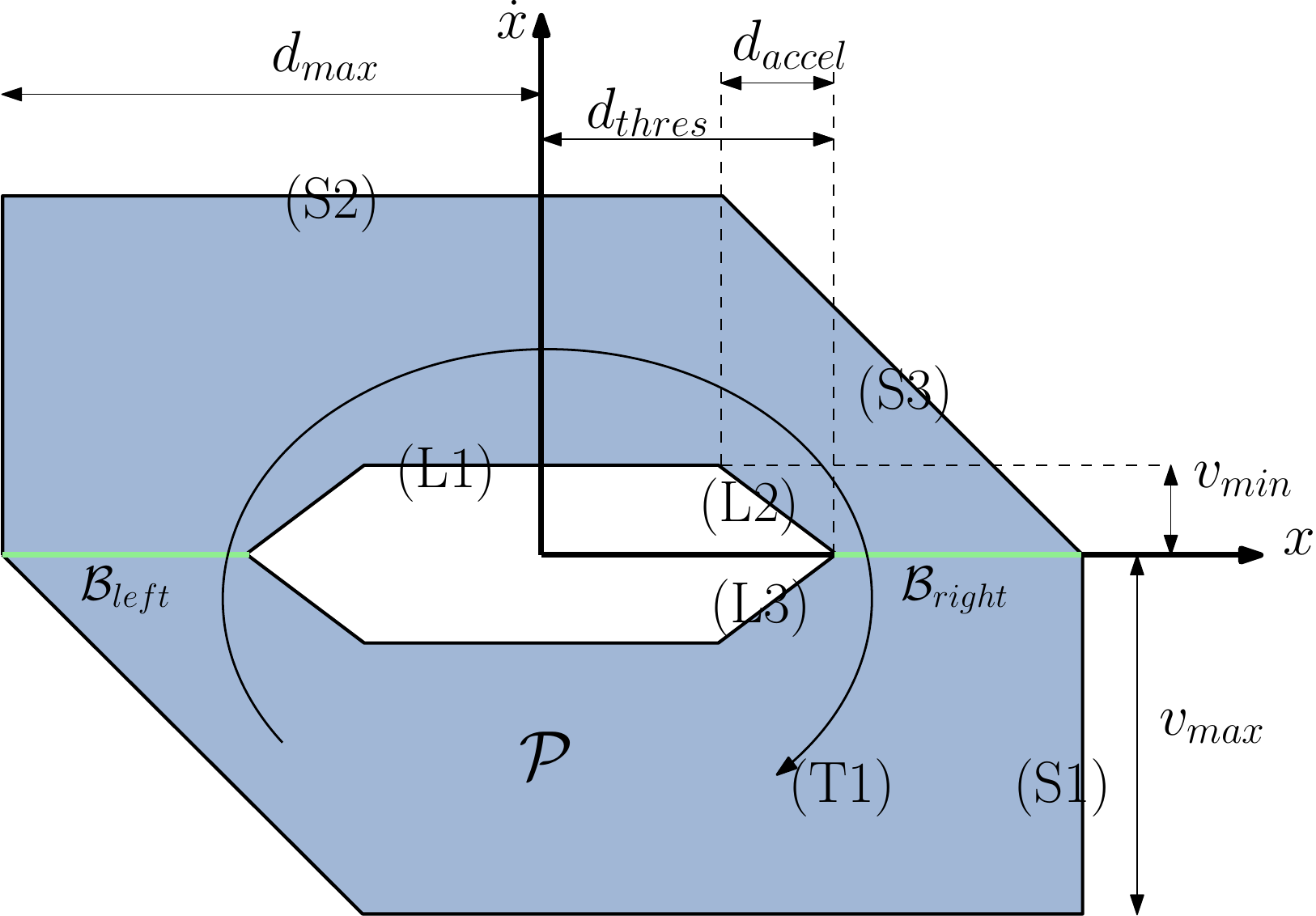}%
\caption{Desired maneuver envelope (in blue). The specifications (S1-3), (L1-3), and (T1) are defined in Section \ref{sec:specs}. The green lines represent the corresponding crossing sets that define when the system switches from the left-to-right to the right-to-left mode and vice versa.}%
\label{fig:polytope}%
\vspace{-2mm}
\end{figure}

In the next step of our methodology, we state the control specifications, which result in a polytope and a sequence of target sets, see Figure \ref{fig:polytope}. 
%During operation, the boundaries of the room are to be strictly adhered to in order to avoid collision. In addition, we specify a maximum speed to ensure the behavior is not too aggressive. Finally, a minimum speed is specified for cruise between the ends of the room. In the $y$- and $z$-directions, the quadrocopter should remain stabilized at some nominal values.

%For simplicity of exposition, the desired side-to-side motion is along the $x$-axis of the inertial frame $\mathcal{O}$ and the yaw angle, $\psi$, is zero. 

%a threshold distance that must be exceeded in order to transition to the opposite mode. 

%Since this maneuver is executed in the $x$-direction, we formulate these specifications in terms of $x$ and $\dot{x}$. Let the middle of the room be at the origin of the inertial frame. We define a set of variables that completely parameterize the maneuver envelope, see Figure \ref{fig:polytope}. For simplicity, the maneuver is symmetric about the origin so that the two modes are oddly symmetric. The distance from the origin to a room boundary is $d_{max}$. The maximum speed is $v_{max}$. The threshold distance as measured from the origin is $d_{thres}$. The minimum cruise speed is $v_{min}$. Finally, a distance parameter $d_{accel}$ is defined to help describe how quickly the quadrocopter should accelerate near the threshold point.

\setcounter{paragraph}{0}
\paragraph{Safety Specifications}
To remain within the boundaries of the room, we require $|x| \leq d_{max}$, see Figure \ref{fig:polytope}. The maximum speed limitation imposes $|\dot{x}| \leq v_{max}$. For a safe turnaround, we impose a deceleration requirement close to the room's boundaries; for compatibility with the RCP approach we use linear inequalities. We define a safe, minimum deceleration, $a_{saf}$, and obtain the following linear inequalities: $|x - \dot{x}/a_{saf} | \leq d_{max}$, where we choose $a_{saf} = -v_{max}/(d_{max}-d_{thres}+d_{accel}) < 0$, see Figure~\ref{fig:polytope}.
For safety to be ensured, all three inequalities must be simultaneously satisfied at all times:
\begin{enumerate} 
\item[{\bf (S1)}] \label{itm:S1}
Position: $|x| \leq d_{max}$.
\item[{\bf (S2)}] \label{itm:S2}
Speed: $|\dot{x}| \leq v_{max}$.
\item[{\bf (S3)}] \label{itm:S3}
Deceleration: $|x - \dot{x}/a_{saf} | \leq d_{max}$.
\end{enumerate}

\paragraph{Liveness Specifications}
%Next we describe the so-called liveness specifications. Their purpose is to make the maneuver more aggressive, so that around the middle of the room the quadrocopter moves with at least the minimum cruise speed. It is most easily visualized as a deleted region centered at the origin, see Figure \ref{fig:polytope}. There are many ways to construct such a region. The only requirement is that it is contained strictly within the safety region described above. 
The liveness specifications ensure that the quadrocopter moves with sufficient speed when cruising between the ends of the room.
Using our parameters as defined in Figure \ref{fig:polytope}, we first define the inequality $|\dot{x}| \geq v_{min}$, which says the speed should be above the minimum. Next we define $|x - \dot{x} / a_{liv} | \geq d_{thres}$, and $|x + \dot{x} / a_{liv} | \geq d_{thres}$, with $a_{liv} = -v_{min}/d_{accel} < 0$. These inequalities describe how the quadrocopter should accelerate from zero speed to the minimum speed and decelerate from the minimum speed to zero speed, respectively. For liveness to be ensured, any of the three inequalities must be satisfied at all times:
\begin{enumerate} 
\item[{\bf (L1)}]
Speed: $|\dot{x}| \geq v_{min}$.
\item[{\bf (L2)}]
Acceleration: $|x - \dot{x} / a_{liv} | \geq d_{thres}$.
\item[{\bf (L3)}]
Deceleration: $|x + \dot{x} / a_{liv} | \geq d_{thres}$.
\end{enumerate}

\paragraph{Desired Temporal Sequence}
We decompose this side-to-side maneuver into two opposing discrete modes: one in which the quadrocopter is moving from left to right, referred to as the L2R; and one in which the quadrocopter is moving from right to left, denoted by R2L. For the L2R and R2L modes respectively, we define target sets to be reached in order to transition to the opposite mode:
\begin{align}
\cB_{right} &= \{(x,\dot{x}) \; | \; x \in [d_{thres}, d_{max}], \; \dot{x}=0\} , \\
\cB_{left} &= \{(x,\dot{x}) \; | \; x \in [-d_{max},-d_{thres}], \; \dot{x}=0\} .
\end{align}
\noindent The sets are shown in Figure \ref{fig:polytope} (green lines). Assuming that the initial mode is L2R,  the sequence of target sets to be crossed by the system trajectory is:
% Also, as in the context of dynamical systems, we refer to the set of points in the state space that comprise a given system's trajectory in time as its \emph{orbit}. Assuming that the initial mode be L2R, the system transitions to the R2L mode when the quadrocopter orbit crosses $\cB_{right}$. Then the system transitions back to the L2R mode when the quadrocopter orbit crosses $\cB_{left}$. This can be repeated indefinitely. The corresponding temporal specification is:
%{\em The quadrocopter is transported to the right side, then to the left side, and back indefinitely}.
\begin{enumerate}  
\item[{\bf (T1)}]
%The quadrocopter's orbit must cross the sets $\cB_{right}$ and $\cB_{left}$ in exactly the following sequence: 
$\cB_{right}$, $\cB_{left}$, $\cB_{right}$, $\cB_{left}$, $\ldots$\,. 
\end{enumerate}

\begin{comment}
\paragraph{Summary}
Collecting all the above inequalities, we can formally describe the operational region (blue shaded region in Figure \ref{fig:polytope}. Since the inequalities are linear, this defines a (non-convex) polytope, denoted as $\cP$. For all the specifications to be satisfied, we require the following logical statement to be true at all times:
\begin{equation} \label{eq:logicspec}
\left[ \text{(S1)} \wedge \text{(S2)} \wedge \text{(S3)} \right] \wedge 
\left[ \text{(L1)} \vee \text{(L2)} \vee \text{(L3)} \right] \wedge \text{(T1)} .
\end{equation}
\end{comment}

For the experiment, we fix the values of the polytope parameters in Figure \ref{fig:polytope} to: $d_{max} = 2.5$ m, $d_{thres} = 1.5$ m, $d_{accel} = 0.3$~m, $v_{max} = 2$ m/s, and $v_{min} = 0.6$ m/s. 

\subsection{Triangulation and Exit Facets}

%To satisfy the specifications from Section \ref{sec:specs}, we use the reach control approach, see \cite{BG14, RB06}, in order to determine a suitable $\theta_d(t) = \tan^{-1}(u(t)/g)$. We construct $u(t) = u(x(t),\dot{x}(t))$ as a feedback using the current $x$- and $\dot{x}$-measurement so that the closed-loop dynamics for the nominal equations of motion in the $(x-\dot{x})$-plane \eqref{eq:ddotxDI} guarantee safety, liveness, and the desired temporal sequence. From Section \ref{sec:specs}, the set $\cP$ determines the allowable $(x,\dot{x})$ states. We partition $\cP$ into a set of regions for which we each specify a controller. Since $\cP$ is a non-convex polytope, we employ a triangulation of $\cP$ into simplices, cf. Figure \ref{fig:triangulation}.

%We present our complete design (including numerical values) for the reach controllers on the polytope $\cP$. To proceed, we fix the values of the motion parameters from Section \ref{sec:specs} to: $d_{max} = 2.5$ m, $d_{thres} = 1.5$ m, $d_{accel} = 0.3$~m, $v_{max} = 2$ m/s, and $v_{min} = 0.6$ m/s. 

\begin{figure}[t]
\centering%
\includegraphics[width=0.9\linewidth,height=0.55\linewidth]{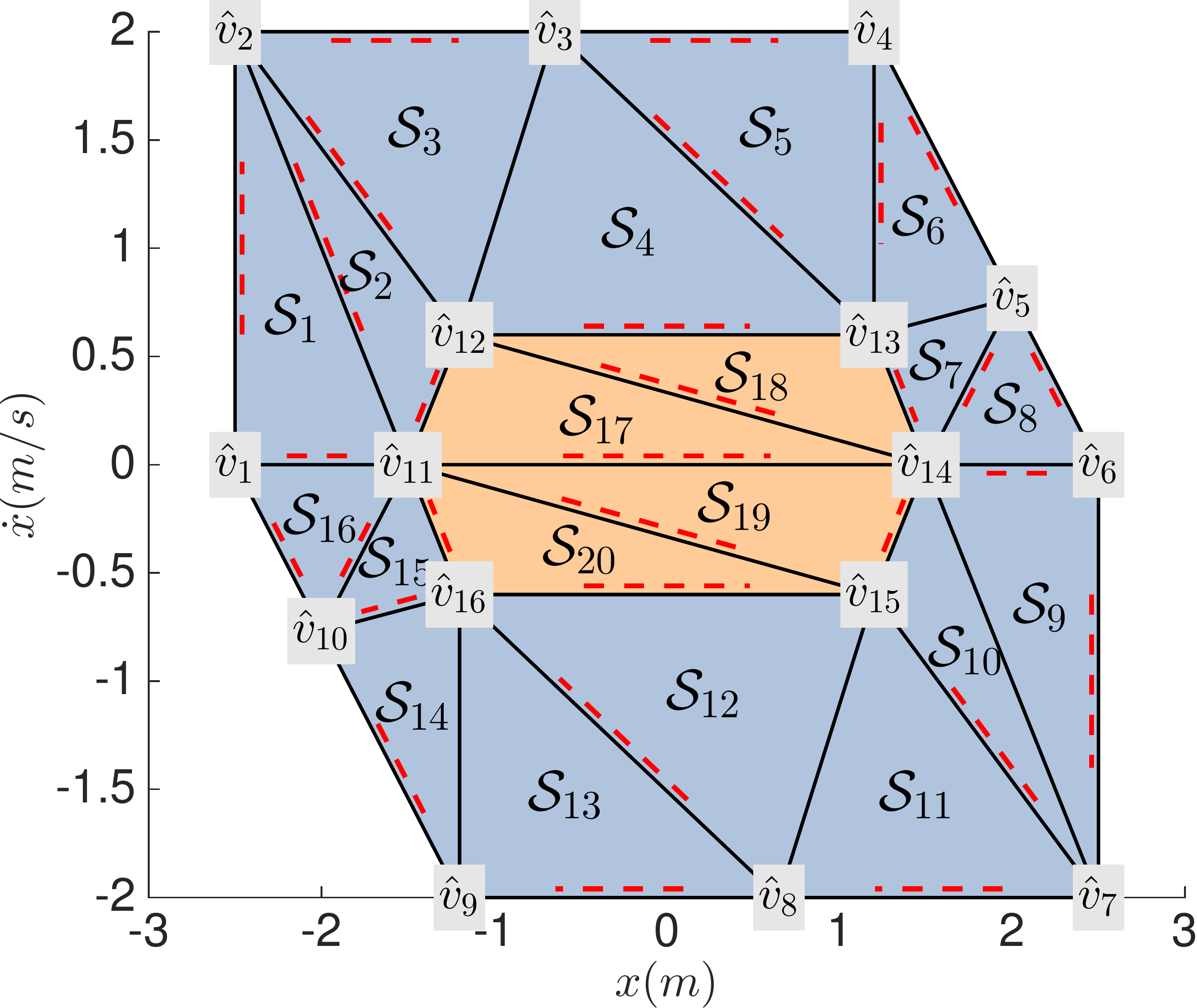}%
\caption{The triangulation of $\cP$, also showing exit facets, for the L2R mode. Red dashed lines represent the restricted  facets of a given simplex, drawn inside of the respective simplex. The blue shaded area represents $\cP$. The orange shaded area is guaranteeing the liveness of the system and must be avoided in nominal operating conditions.}%
\vspace{-2mm}
\label{fig:triangulation}%
\end{figure}

Our triangulation is shown in Figure \ref{fig:triangulation}. The vertices of the triangulation are uniquely labeled as $\hat{v}_i$, $i \in \mathcal{I}_v := \{1, \ldots, 16\}$.
The simplices are uniquely labeled as $\cS_i$, $i \in \mathcal{I}_s := \{1, \ldots, 20 \}$. Although automated procedures for triangulating and solving RCP in conjunction are considered in~\cite{Bel08}, here the triangulation was naively generated by manually partitioning $\cP$ into simplices. 

Next, to define the sequence of simplices to be visited, for each simplex we choose its restricted and exit facets. This information is also encoded in Figure \ref{fig:triangulation} via the red dashed lines. For the L2R mode, the facets were chosen so as to ensure that the resulting closed-loop vector field causes trajectories to reach the set $\cB_{right}$. Due to symmetry, the R2L can be implemented trivially using the L2R mode's design.

We remark that we also triangulated the non-liveness region (the orange region in Figure \ref{fig:triangulation}) in order to improve the robustness of our design. In the case that the system enters this region due to a large disturbance, the controllers defined on $\cS_i$, $i = \{17,18,19,20\}$, return the system to nominal behavior.

\subsection{Resulting Controller Design}

Now we solve RCP on each simplex $\cS_i$, $i \in \mathcal{I}_s$. Equation~\eqref{eq:ddotxDI} showed that under perfect stabilization in the $y$- and $z$-directions, the dynamics in the $x$-direction reduce to $\ddot{x} = u$. For this double-integrator model, the dynamics \eqref{eq:affinesys} over each simplex $\cS_i$ are 
\begin{equation}
\dot{s} = As+Bu+a= \begin{bmatrix}0 & 1 \\ 0 & 0 \end{bmatrix}s + \begin{bmatrix} 0 \\ 1 \end{bmatrix}u ,
\end{equation}
\noindent where $s := (x,\dot{x})$, whose components are the $x$-position and $x$-velocity of the quadrocopter, respectively. 
%Clearly we apply the formulas of Section \ref{sec:RCPmath} with $n = 2$.

With the triangulation and exit facets presented, we now construct the controllers on each simplex. For each simplex~$\cS_i$, $i \in \mathcal{I}_s$, we identify the three corresponding vertices $v_j = \hat{v}_{i_j}$, $j \in \mathcal{I}_2$, $i_j \in \mathcal{I}_v$, that form the simplex along with the corresponding exit facets $\mathcal{I}_r$. We design the control values at the three vertices such that the invariance conditions~\eqref{eq:invconds} are satisfied. Note that larger control values result in more aggressive controls, and hence actuator constraints can also be incorporated. Finally, we use \eqref{eq:affinefb} to obtain the feedback law \eqref{eq:affinefblaw}. The controllers can be computed manually or using a numerical solver.
%The point of equilibria is handled later.
%PUT THIS SOMEWHERE: For our double integrator model, it is straightforward as equilibria can only occur along the $x$-axis.

\begin{figure}[t]%
\centering%
\includegraphics[width=0.9\linewidth,height=0.55\linewidth]{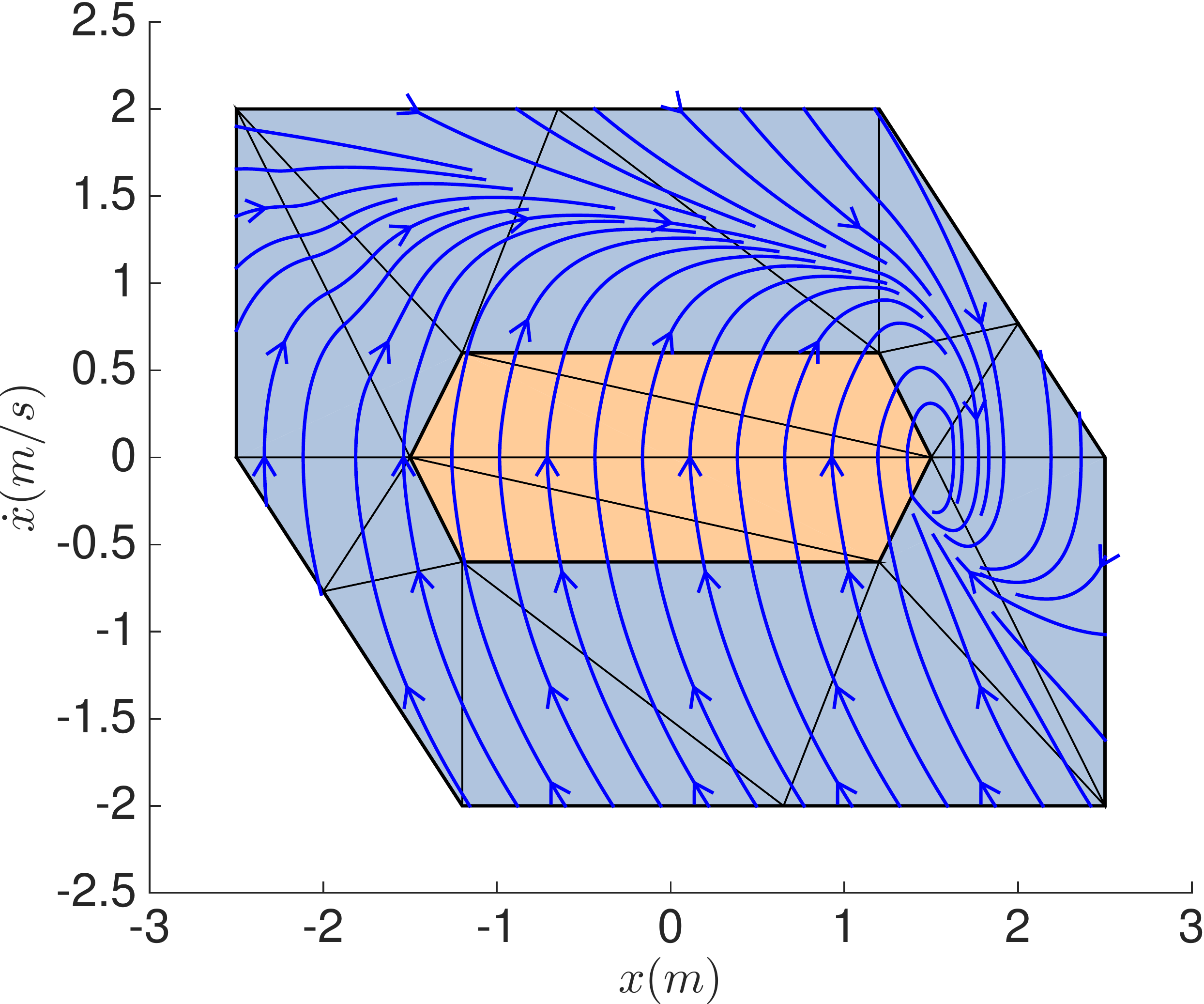}%
\caption{The closed-loop vector field for the L2R mode, illustrated by the direction of the blue arrows.}%
\vspace{-3mm}
\label{fig:polytopeVF}%
\end{figure}

%The resulting control values $\hat{u}_{i_j} $, $i_j \in \mathcal{I}(\hat{v}_i)$, for our specific example are listed in Table \ref{tbl:ctrvals}. The actual feedback over each simplex can be constructed using \eqref{eq:affinefb}, and the corresponding vertex and control values. A plot of t
The resulting closed-loop dynamics for the L2R mode over~$\cP$ and the non-liveness region is shown in Figure \ref{fig:polytopeVF}. As expected, trajectories above the $x$-axis that start in $\cP$ remain inside $\cP$ and are guided  towards the set $\cB_{right}$, while any other non-nominal trajectory (starting in $\cP$ below the $x$-axis or in the non-liveness region) eventually recovers and crosses $\cB_{right}$.
The closed-loop vector field for the R2L mode is obtained by oddly reflecting the L2R design about the origin.\begin{comment}
\begin{table}
\centering
\begin{footnotesize}
\caption{Determined Control Values}
\label{tbl:ctrvals}
\begin{tabular}{| c | l | c | l |}
\hline
Vertex & Control value & Vertex & Control value\\
\hline 
$\hat{v}_1$ & $3.1421$ & $\hat{v}_2$ & $-0.5383$ \\
\hline
$\hat{v}_3$ & $-1.5135$ & $\hat{v}_4$ & $-3.1421$\\
\hline 
$\hat{v}_5$ & $-3.1421$ & $\hat{v}_6$ & $-3.1421$\\
\hline
$\hat{v}_7$ & $3.1421$ & $\hat{v}_8$ & $3.1421$ \\
\hline
$\hat{v}_9$ & $3.1421$ & $\hat{v}_{10}$ & $3.1421$\\
\hline 
$\hat{v}_{11}$ & $1.3844$ & $\hat{v}_{12}$ & $2.4170$\\
\hline
$\hat{v}_{13}$ & $1.0474$ & - & - \\
\hline
$\hat{v}_{14}$ & $0.3980$ & $\hat{v}_{14}$ & $-1.3844$   \\
& for $ \cS_{7}, \cS_{17}, \cS_{18}, \cS_{19}$ & & for $ \cS_{8}, \cS_{9}, \cS_{10}$ \\
\hline
$\hat{v}_{15}$ & $3.1421$ & $\hat{v}_{16}$ & $3.1421$ \\
\hline
\end{tabular}
\end{footnotesize}
\end{table}
\end{comment}

We remark that to avoid discontinuities in the control when transitioning between simplices, we match the control values at the vertices along shared facets between contiguous simplices. For technical reasons related to solving RCP, it is necessary to introduce a discontinuity at $\hat{v}_{14}$ \cite{BG14}. 

\section{Experimental Results}
\label{sec:experiment}

\begin{figure}[t]%
\centering%
\includegraphics[width=0.8\linewidth,height=0.55\linewidth]{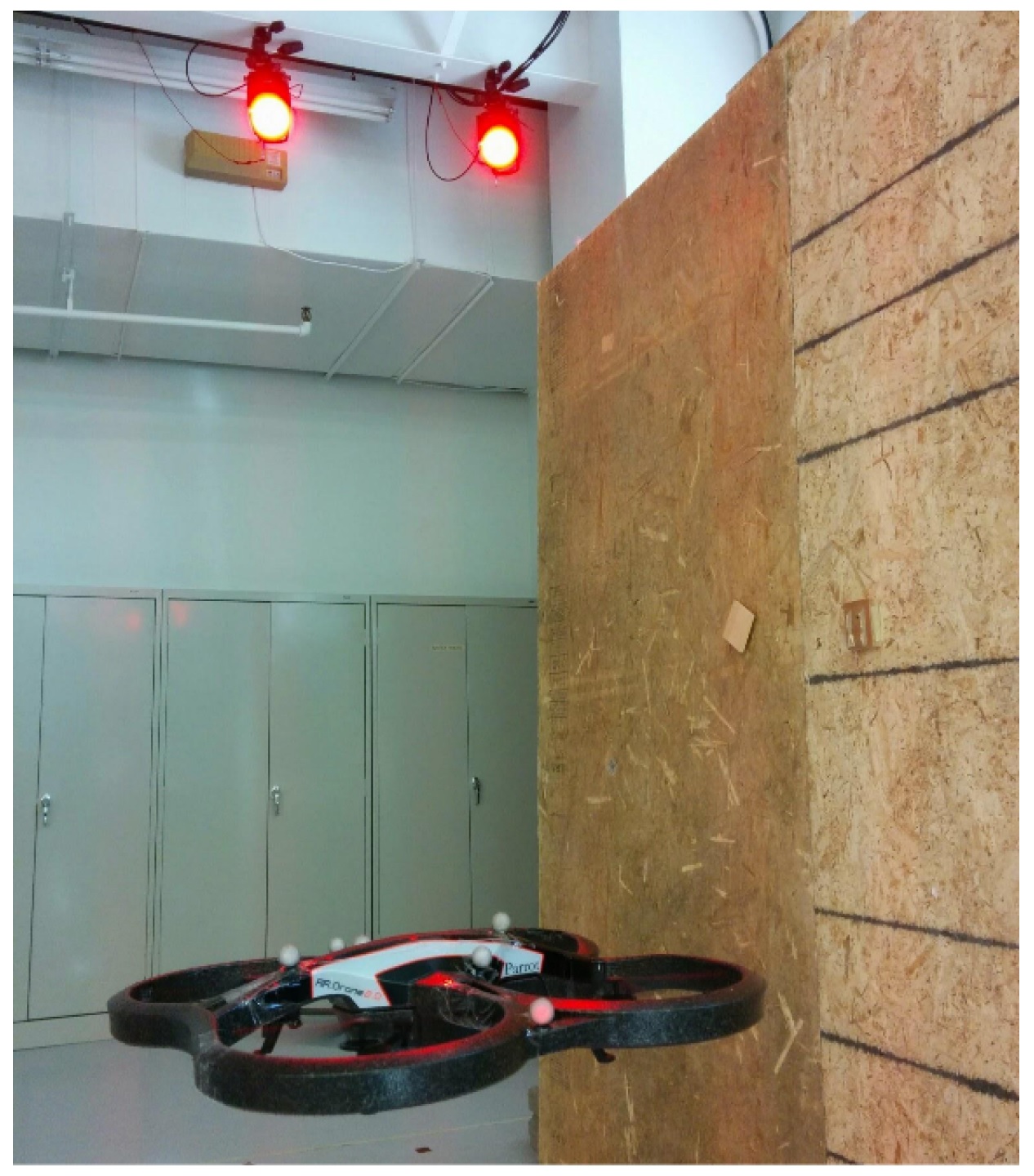}%
\caption{Our quadrocopter vehicle close to the wall of the room with the motion capture camera system in the background.}%
\vspace{-2mm}
\label{fig:expsetup}%
\end{figure}

Our experimental platform is the Parrot AR.Drone 2.0 running firmware version 2.3.3. We interface with the AR.Drone through ROS, an open-source robot operating system \cite{ARDRONE}.
More precisely, we used ROS Hydro, installed on a 64-bit 12.04 Ubuntu version. In addition, we used the ROS \emph{ardrone autonomy} package \cite{ARDRONE}, version 1.3.1. All experiments were conducted with the indoor hull shown in Figure \ref{fig:expsetup}, which protects the vehicle propellers.

%We first demonstrate the success of both a standard tracking method and our RCP approach to execute the desired side-to-side motion under nominal conditions. 
We demonstrate the successful execution of the desired side-to-side motion based on our RCP approach and compare it to the performance of a standard trajectory tracking controller \cite{SCH2011,LUP2014}, which guides the vehicle along a predefined, timed side-to-side trajectory. 
%Then to illustrate the robustness of our method, we introduce a disturbance not accounted for in the control design. We show that our control strategy still transitions between the two discrete modes correctly while the standard tracking control approach fails to do so. 
A video showing the experimental results can be found at: \url{http://tiny.cc/quadrotorRCPx}.

%The tracking controller for starting at the left side in the L2R mode is based on tracking the following signal in the $x$-direction:
%\begin{equation} \label{eq:trackCos}
%x_d(t) = 2 \cos (0.65 t + \pi) .
%\end{equation}
%For appropriately chosen amplitude $A$ and frequency $\Omega$,
%\noindent The resulting orbit $(x_d(t),\dot{x}_d(t))$, which is an ellipse, fits inside the polytope $\cP$ for all time and executes the desired temporal sequence. The switches between modes occur automatically whenever the velocity $\dot{x}_d(t)$ crosses 0. The standard controller computes the desired pitch $\theta_d(t)$ based on the desired position $x_d(t)$ and the measured state of the vehicle and replaces the reach controller in Figure \ref{fig:architecture}. 

\begin{figure}[t]
\centering%
\includegraphics[width=1\linewidth,trim=2cm 0cm 3.5cm 0.5cm, clip=true]{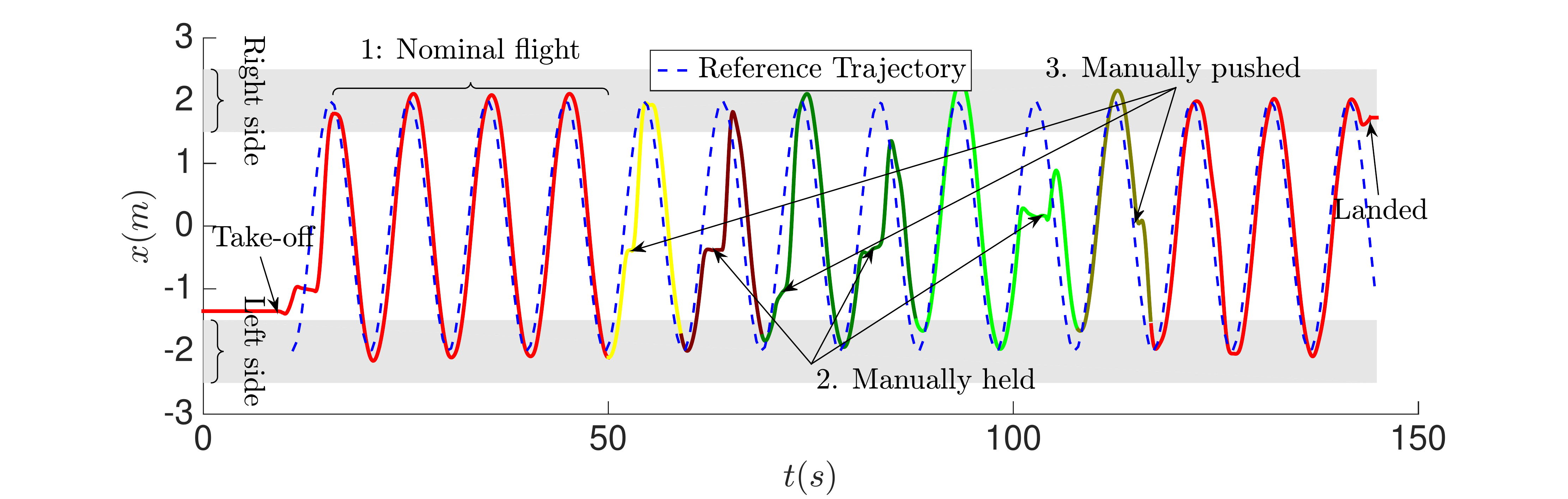} \\
\includegraphics[width=1\linewidth,trim=5cm 1cm 5cm 1.5cm, clip=true]{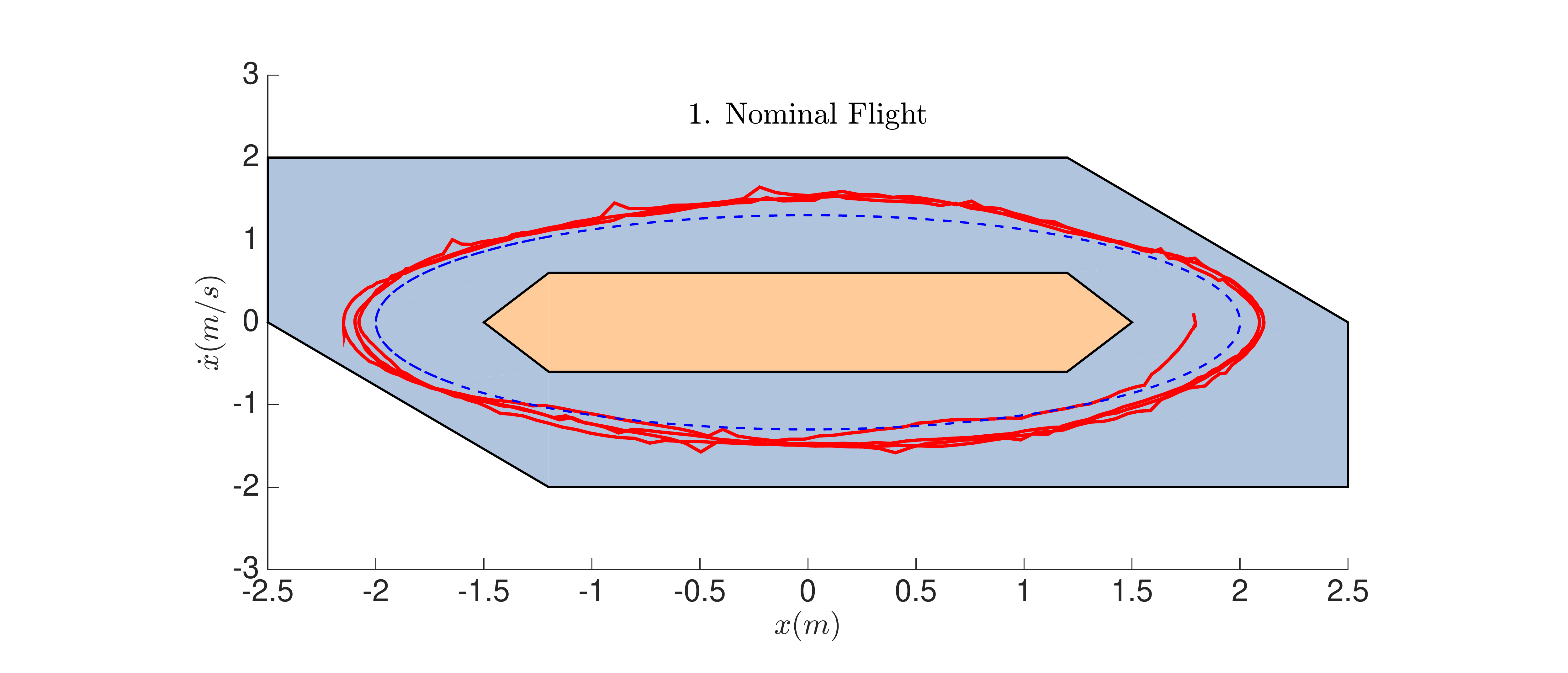} \\
\includegraphics[width=1\linewidth,trim=5cm 1cm 5cm 1cm, clip=true]{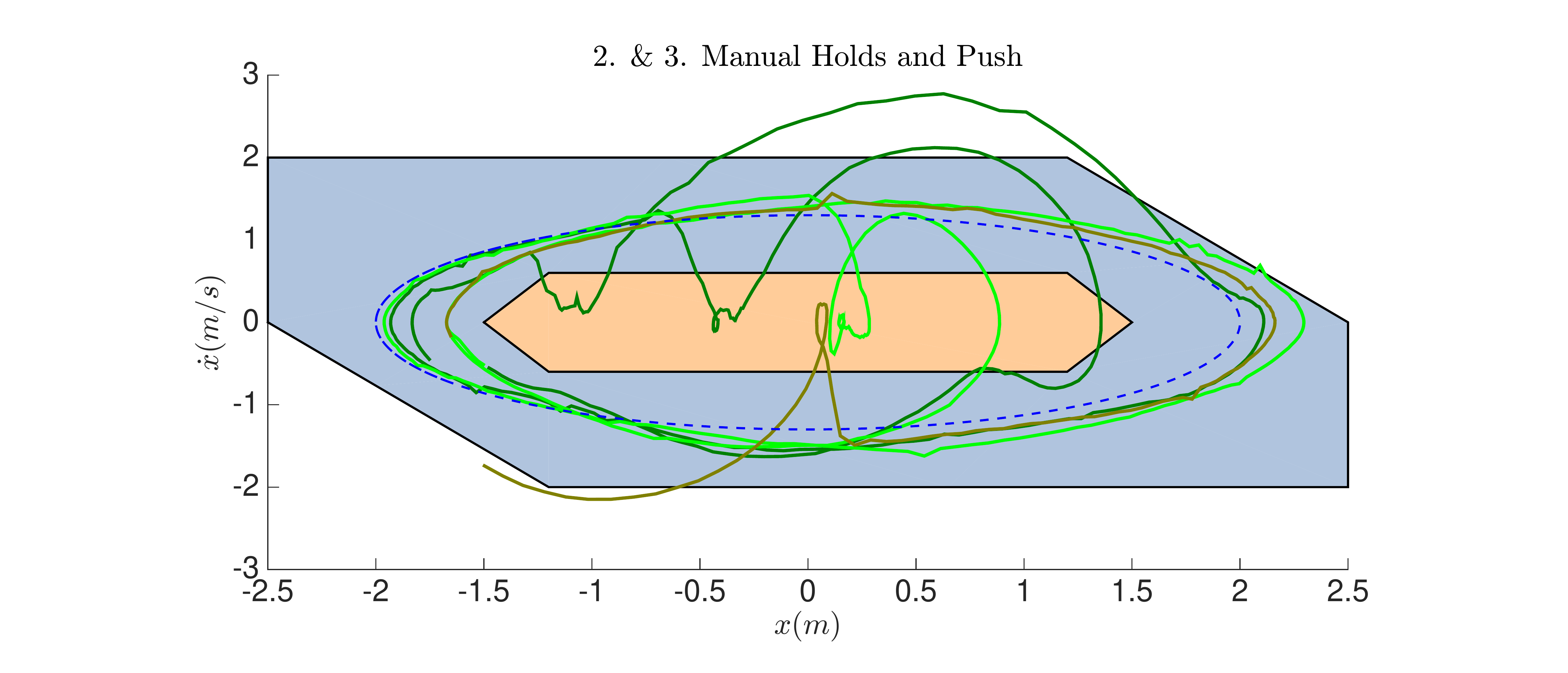} \\
\caption{Experimental results of the tracking approach.}%
\vspace{-2mm}
\label{fig:trackexp}%
\end{figure}

\begin{figure}[t]
\centering%
\includegraphics[width=1\linewidth,trim=2cm 0cm 6cm 0.5cm, clip=true]{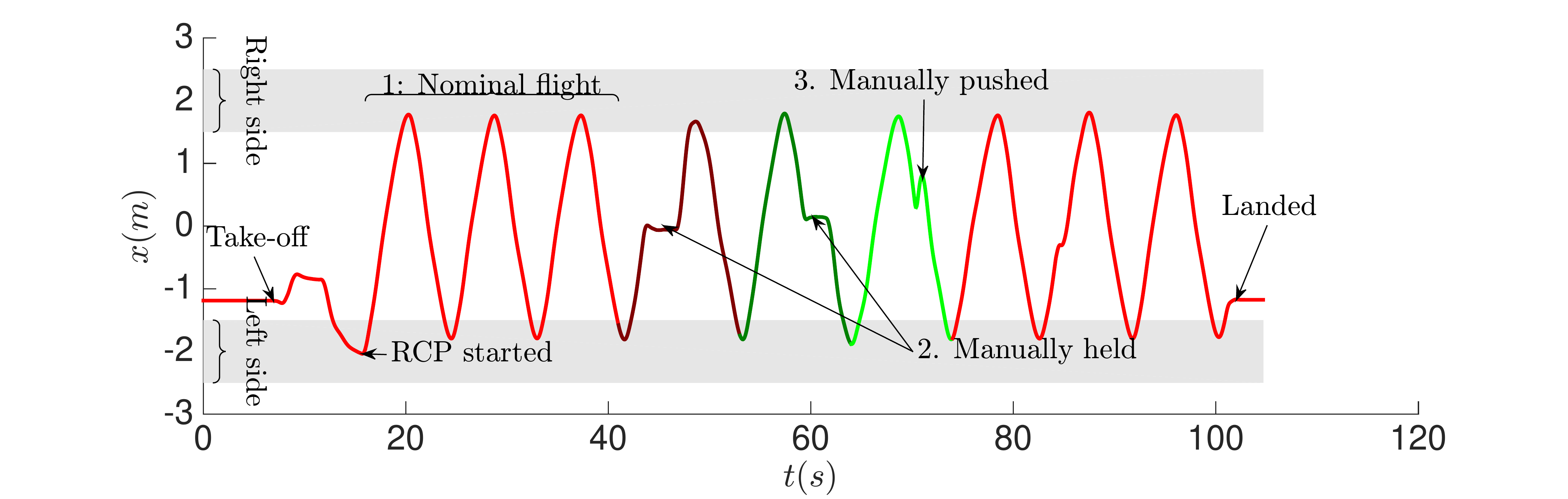} \\
\includegraphics[width=1\linewidth,trim=5cm 1cm 5cm 2cm, clip=true]{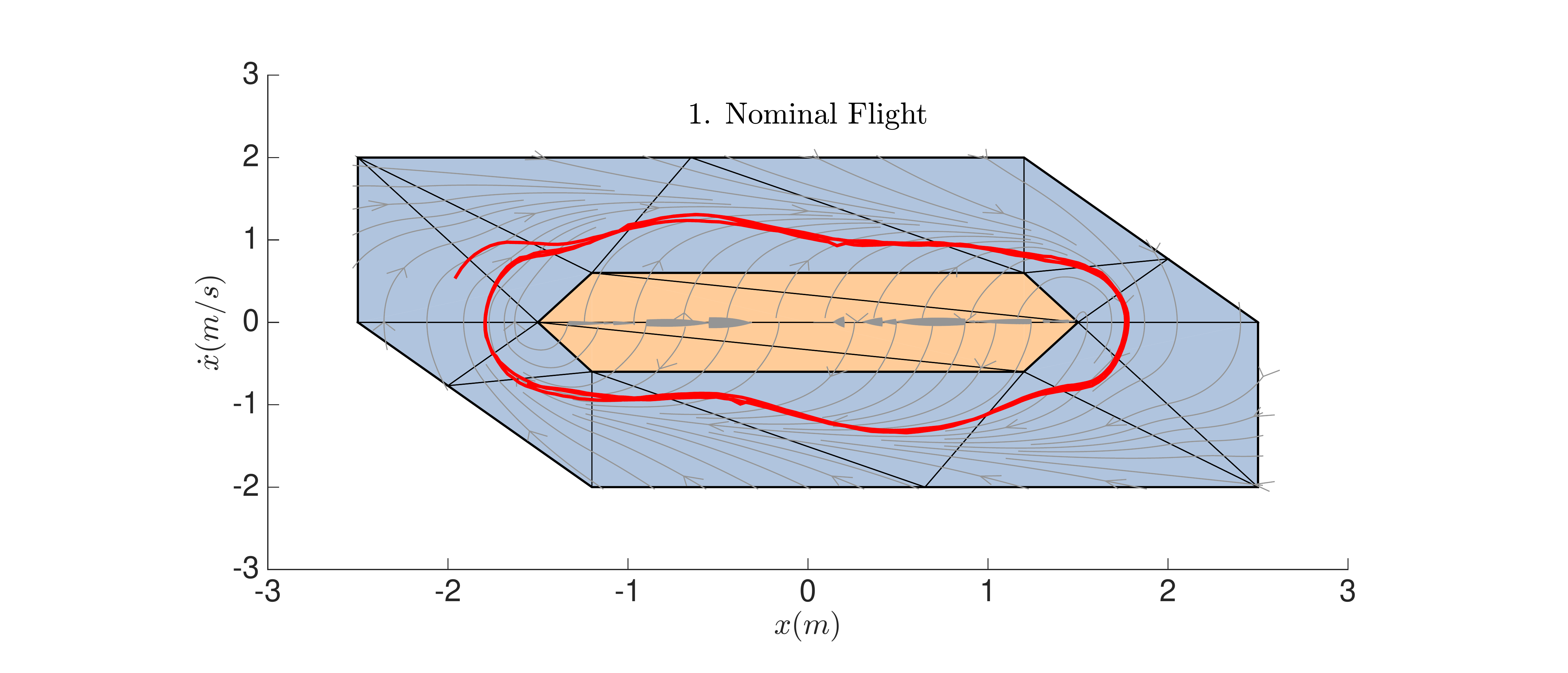} \\
\includegraphics[width=1\linewidth,trim=5cm 1cm 5cm 2cm, clip=true]{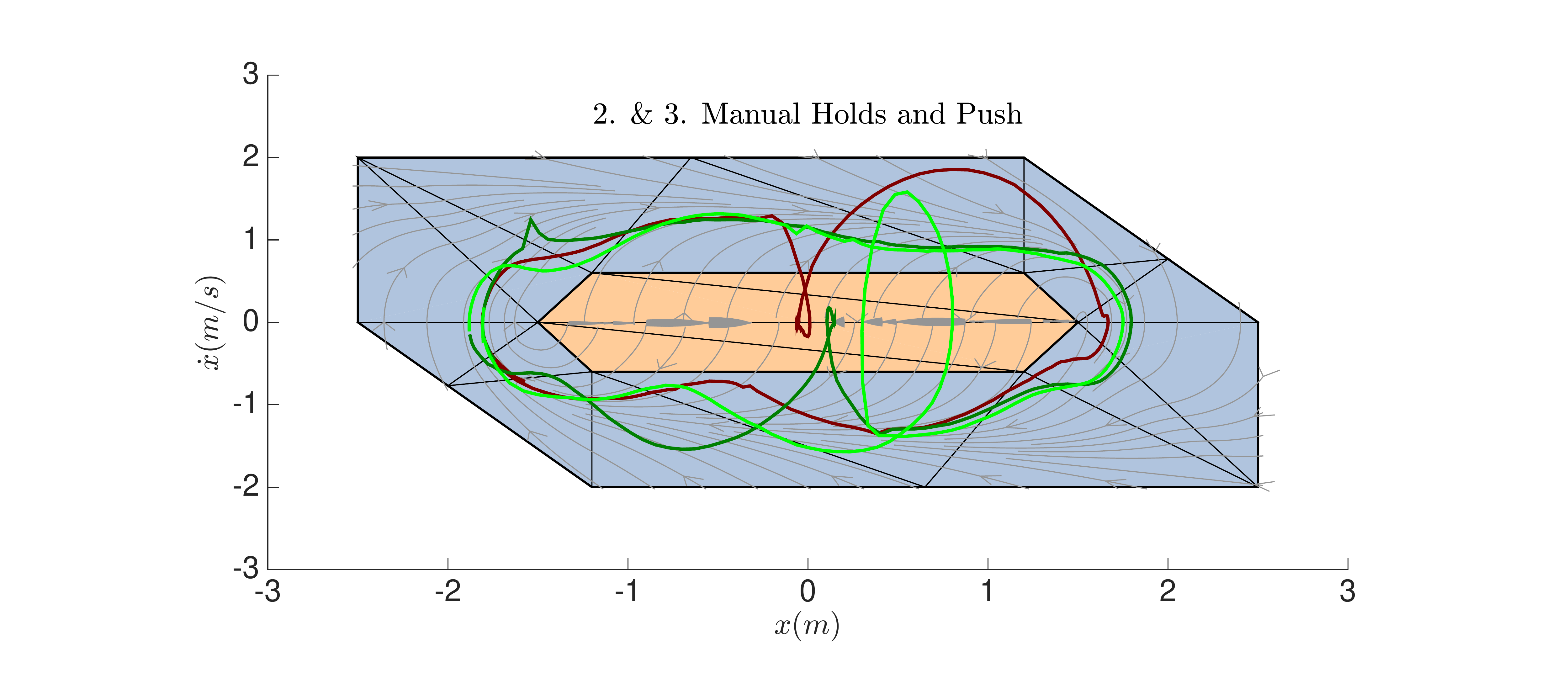} \\
\caption{Experimental results of the RCP approach.}%
\vspace{-2mm}
\label{fig:rcpexp}%
\end{figure}

In our experiments, the following actions were performed for both trajectory tracking and the RCP approach: 1) nominal flight consisting of a few cycles of the L2R and R2L modes, 2) introducing a disturbance by manually holding the vehicle, and 3) introducing a disturbance by pushing the vehicle. At the top of Figures \ref{fig:trackexp} and \ref{fig:rcpexp}, we show the position $x(t)$ over time. The key difference that we observe is that when the quadrocopter is disturbed, the tracking approach fails the desired temporal sequence whereas the RCP approach does not. The middle plots show that the nominal behavior of both methods are comparable. The bottom plots show the significant degradation caused by the disturbances, where only in the tracking approach the trajectories exit the safety region by speeding up too much. 

\section{Conclusion}
We succeeded in experimentally demonstrating the first ever implementation of reach controllers on a real system. The result is a logically complex quadrocopter maneuver. The main advantages of the reach control approach are that it permits the incorporation of safety constraints, event sequences and logical constraints, and robustness to unmodeled disturbances, as shown in our comparison between the reach control and standard tracking approaches. The side-to-side maneuver shown here was mainly chosen to demonstrate the proof-of-concept, and so an extension would be to apply the RCP methodology to a more complex maneuver and compare to path following controllers.
%A first extension would be to incorporate reach control also in the $y$-direction or, by yawing the vehicle, use a hybrid-based framework to switch between oriented side-to-side motions to achieve more complex motions. Another extension is to explore the possibility of eliminating the remaining standard tracking controllers with appropriate reach controllers to be able to logically control all the degrees of freedom of the quadrocopter.

\bibliographystyle{IEEEtranS}

\end{document}